# IMECE2021-69936

# DESIGN OPTIMIZATION OF MONOBLADE AUTOROTATING PODS TO EXHIBIT AN UNCONVENTIONAL DESCENT TECHNIQUE USING GLAUERT'S MODELING.

Shashwat Patnaik[1], Kanishk[1]

[1]Delhi Technological University, New Delhi, India

## ABSTRACT

*Many unconventional descent mechanisms are evolved in nature to maximize the dispersion of seeds to increase the population of floral species. The induced autorotation produces lift through asymmetrical weight distribution, increasing the fall duration and giving the seed extra time to get drifted away by the wind. The proposed bio-inspired concept was used to produce novel modern pods for various aerospace applications that require free-falling or controlled velocity descent in planetary or interplanetary missions without relying on traditional techniques such as propulsion-based descent and the use of parachutes. We provide an explanation for the design procedure and the functioning of a mono blade auto-rotating wing. An element-based computational method based on Glauert's blade element momentum theory (BEMT) model was employed to estimate the geometry by maximizing the coefficient of power through MATLAB's optimization toolbox using the Sequential quadratic programming (SQP) solver. The dynamic model was developed for the single-wing design through the MATLAB Simulink 6-DOF toolbox to carry out a free-flight simulation of the wing to verify its global stability.*

## 1. INTRODUCTION

Blade Element Momentum Theory (BEMT) presents a model to describe the characteristics of a turbine blade according to its mechanical, geometric parameters and characteristics of the interacting flow [11]. BEMT combines two theories; Blade Element Theory proposed by William Froude in 1878 and the Momentum Theory proposed by William Rankine in 1865. Blade Element Theory (BET) cuts the turbine's blade into sections, also called wing's elements, and each of them is being approximated by a planar model, where it studies the performance of each elements from a local point of view [1]. Blade Element Theory is used to obtain aerodynamic forces being applied onto the individual wing's element as functions of the blade geometry and flow characteristics. However, the momentum theory is a global theory and models the behavior of the channel of fluid traversing through the turbine's blade by assuming a macroscopic point of view [2].

In 1926, Hermann Glauert proposed the BEMT or Blade Element Momentum Theory, which incorporates both theories; Blade Element Theory and Momentum Theory. Moreover, he improved the Momentum theory by accounting for the rotation of the fluid rings caused by the turbine's interaction with the fluid flow [1]. Glauert's theory aims to formulate algebraic functions to model the relationship between a flow and a rotating blade. Gluaert's model characterizes the interaction using two descriptions; and a local model which describes the 2D behavior of a wing's element in corresponds to fluid flow and a global model that summarizes the progression of fluid rings flowing across the turbine's blade [2].

### 1.1 Literature Review

Bio-inspired Micro Air Vehicles have been gaining popularity in recent years due to their mechanical simplicity and compact design [10]. Many studies have been done to achieve stable, decent flights and directional flights in most indoor and limited environments. Few papers examined the flight capability of single-wing autorotating design inspired by maple seed descent mechanism.

The paper by Jeremy Ledoux outlines a framework for a model based on H. Glauert [1]. It models the aerodynamic interaction between the blade of a turbine and a fluid flow across it. It proposed a BEM equation that reformulates Glauert's model, establishing the criteria for which solution will always exist. It also explores various algorithms for the convergence of the solution.

The author Shane Win proposed a lightweight payload aerial deployment module to deliver to disaster-struck areas in his paper [3]. It was based on a samara autorotating wing; however, it had multiple individual units attached to form a

 

collective rotor hub. The rotor had a simple release mechanism to enable individual wings to separate to adapt to different mission requirements. Each blade has a control surface whose angle can be changed to reach its required destination. Moreover, the author conducted experiments using a motion capture system to evaluate the collective rotor hub and individual wings control performance in real-world conditions.

Shane Kyi Hal Win's paper presents a chordwise optimization for a single wing in steady-state autorotative flight [4]. An actuator is present in the wing to control the descent's direction by changing its angle of attack. Further, they performed numerical analysis and vertical wind tunnel simulation to verify the design's stability under various conditions.

The paper by Kapil Varshney investigates how a maple seed reaches its steady-state helical autorotation motion using a high-speed digital camera [5]. They discovered that aerodynamic forces are not entirely responsible for its autorotation. The authors identified three essential parameters which dictate its helical motion. They determined that the coupling between aerodynamic torque and rigid-body dynamics separates the maple seed's flight pattern from various other autorotation cases.

All of the examined papers and articles failed to produce a comprehensive and well explained process of calculating and estimating the wing shape parameters. This paper aims at filling that void in the knowledge.

**1.2 Free Body Diagram**

The maple seed accomplishes its autorotating state due to three essential factors. One of them is asymmetricity in the mass distribution, i.e. center of mass is present far away from seed's centroid. Another reason is aerodynamic damping, which initiates a tilt of the seed, leading to its helical motion due to rigid-body dynamics. The last factor is the aerodynamic forces, which are responsible for balancing the weight of the seed and centrifugal force [5].

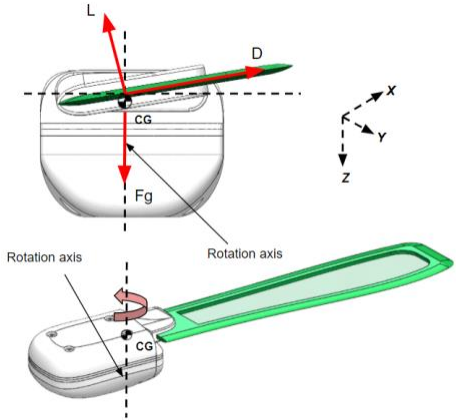

**FIGURE 1:** FREE BODY DIAGRAM OF MONO-BLADE POD

The *figure 1*, illustrates the free-body diagram of the proposed mono-blade pod design. Inertial coordinate *XYZ* is assumed. The origin of the inertial coordinate frame coincides with the center of mass of the mono-blade pod. Moreover, the gravitational force ($F_g$) aligns with the *z-axis* in the same direction such that the gravitational acceleration is positive. The angle of attack of the wing is represented by $\varphi$. Due to the autorotation of the wing, the lift $(L)$ is generated perpendicular to the wing. Further, drag $(D)$ is generated along the wing. Lift and drag forces of the wing generate the cumulative normal and axial forces [8]. The position of the center of mass of the mono-blade pod with respect to the *XYZ* coordinate is represented by $p = (p_x, p_y, p_z)$, the angular position is represented by $\emptyset = (\varphi, \theta, \psi)$. Similarly, its translation velocity is represented by $v = (v_x, v_y, v_z)$ and angular velocity $\omega = (\omega_x, \omega_y, \omega_z)$ with respect to XYZ coordinate system [8].

## 2. OPTIMIZATION FOR WING'S CHORD FUNCTION

The *Figure 1* shows the prototype of the design, a single wing autorotating pod. The design primarily consists of a body (to house the electronics) and a wing. The design provides a simple mechanism to replace the wing and change the wing's angle of attack. The wing is a critical component as it produces the majority of the aerodynamic forces necessary for the autorotation of the mono-blade pod. The mono-blade pod design has a specified footprint, limiting the length and the area of the wing. Hence, the planform or chordwise optimization of the wing to achieve a maximum coefficient of power is quintessential. Since the scale of the mono-blade pod is relatively small, airfoil and wing twist and chamber were not considered for the optimization; to simplify the fabrication process. Instead, the wing was considered to be a flat plate of small thickness.

**2.1 Glauert's Model**

Glauert's model is implemented due to the similarities between the pod's wing and turbine blade interacting with the airflow [1]. The model is derived from Blade Element Momentum Theory. It divides the wing into multiple sections, which generates its own aerodynamic forces. Glauert's establish a relationship that characterizes the interaction between a flow and a section of the rotating wing.

The flow is assumed to be incompressible, horizontal and the air-stream is steady, which implies that the left and right neighborhood of the wing has the same velocity $U_0$. Moreover, $U_{-\infty}$ and $U_{+\infty}$ is denoted as the upstream and downstream flow velocities, respectively [1]. Glauert's Model disregards the interaction between consequent wing elements. Nevertheless, only one section of the wing is considered and assumed to have a constant rotation speed $\Omega$. A fixed value parameter is taken:

$$\lambda = \frac{\Omega r}{U_{-\infty}} \quad (1)$$

Where $'r'$ represents the distance between the specific wing's element to the rotation axis, which in this model is assumed to be at the center of gravity of the pod. As $\Omega$ is assumed to be constant in steady-state, $\lambda$ is only a function of $'r'$ i.e. $\lambda = f(r)$. Hence, the variable $\lambda$ is used to describe the location of the wing's element with respect to the axis of rotation.

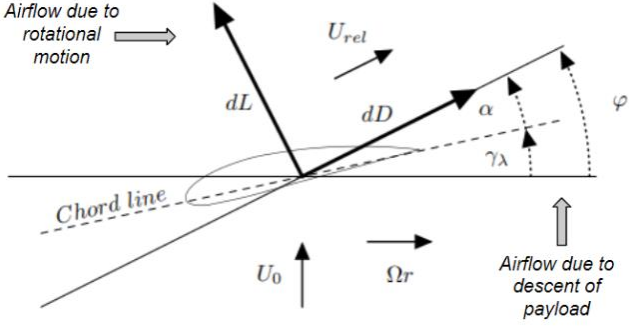

**FIGURE 2:** FORCES ON WING'S ELEMENT DURING AUTOROTATION

Glauert's Model presents the system for each wing's element by relating the variables $a, a'$ and $\varphi$ associated with a ring of fluid that interacts with the element of the wing [2]. The variable $a$ denotes the axial induction factor and defined by:

$$a = \frac{U_{-\infty} - U_0}{U_{-\infty}} \quad (2)$$

The variable $a'$ denotes the angular induction factor and defined by:

$$a' = \frac{\omega}{2\Omega} \quad (3)$$

Where, $\omega$ denotes the rotation speed of the ring of fluid. The variable $\varphi$ is the relative angle of deviation of the ring [12]. It is defined by:

$$\tan \varphi = \frac{1-a}{\lambda(1-a')} \quad (4)$$

The relative speed or the apparent fluid speed is represented by $U_{rel}$ perceived from the wing's element. Relative speed can be represented using the definition of $\varphi$, we get:

$$U_{rel} = \frac{U_0}{\sin \varphi} \quad (5)$$

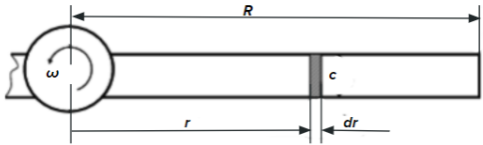

**FIGURE 3:** ILLUSTRATION OF WING ELEMENT FOR POD

The aerodynamic forces on the wing's element, for a given lift coefficients $C_L$ and drag coefficient $C_D$ are given as:

$$dL = C_L(\alpha) \frac{1}{2} \rho \, U_{rel} \, c_\lambda(r) \, dr \quad (6)$$
$$dD = C_D(\alpha) \frac{1}{2} \rho \, U_{rel} \, c_\lambda(r) \, dr \quad (7)$$

Where, $\rho$ denotes the density of the fluid, and the variable $\alpha$ is the effective angle of attack which is the angle between the chord and the direction of flow i.e. $\alpha = \varphi - \gamma\lambda$, where, $\gamma\lambda$ is the twist or local pitch angle. Components, $dD$, and $dL$ are the elementary drag and lift forces generated by wing's element of thickness $dr$ and has a chord represented by the function $c_\lambda(r)$. It is assumed that the coefficients $C_L$ and $C_D$ are only a function of $\alpha$ because the flow is assumed to be low Reynold number [9]. In our model, flat plate airfoil is implemented; hence the function for $C_L$ and $C_D$ are obtained from paper by Jones [6].

$$C_D = \frac{2\pi \sin^2 \alpha}{4 + \pi \sin \alpha} \quad (8)$$

$$C_L = \frac{2\pi \sin \alpha \cos \alpha}{4 + \pi \sin \alpha} \quad (9)$$

In the paper by Varshney, tip losses are indicated to affect the maple seed's autorotation significantly. Glauert's Momentum Theory assumed that the forces on the fluid flow from the wing are constant in each annular element. Though, that represents a rotor system with an infinite number of wings. However, for the system like maple seed, where there is only one wing, Glauert further modified the BMET theory to include the flow over the tip of the wing. It implemented the circulation of the fluid, which exponentially decreases to zero along the length of the wing i.e. $r \to R$. Glauert used the Prandtl Tip Function denoted as $F_\lambda$ [13].

$$F_\lambda(\varphi) = \frac{2}{\pi} \cos^{-1}\left( \exp\left( -\frac{\frac{B}{2}\left(1 - \frac{r}{R}\right)}{\left(\frac{r}{R}\right) \sin \varphi} \right) \right) \quad (10)$$

Where, $B$ represents the number of wings in the system, $R$ is the length of the wing, and $r$ is the distance of the wing's element from the rotation axis.

Glauert's Model of BEMT provides a method to evaluate and optimize the efficiency and the performance of the given geometry rotor system. It provides a function to optimize the $\gamma\lambda$ and $c_\lambda(r)$. Through, Glauert's Model, we can optimize the chord function for a given wing to acquire a desired coefficient of power $C_p$, which corresponds to the ratio of received energy to the captured energy. In the paper by Ledoux, presents function for coefficient of power $C_p$ deceived from Glauert's Model [1].

$$C_p(\gamma\lambda, c_\lambda(r), \varphi, a, a') = \frac{8}{\lambda_{max}^2} \int_{\lambda_{min}}^{\lambda_{max}} \lambda^3 J_\lambda(\gamma\lambda, c_\lambda(r), \varphi, a, a') \, d\lambda \quad (11)$$

Where, $J_\lambda$ is:

$$J_\lambda(\gamma\lambda, c_\lambda(r), \varphi, a, a') = F_\lambda(\varphi) \, a'\,(1-a)\left(1 - \frac{C_D(\varphi - \gamma\lambda)}{C_L(\varphi - \gamma\lambda)} \tan^{-1}\varphi\right) \quad (12)$$

 

## 2.2 Optimization

It is quintessential to decrease the descent velocity of the pod below its terminal speed. The function for the coefficient of power obtained from Glauert's Model gives the relationship between the chord function $c_\lambda(r)$ and coefficient of power. The optimization to maximize the coefficient of power $C_p$ will yield a wing's planform design or chord at each $r$. It is assumed that the airflow is in a steady-state, and the pod is descending into undisturbed clean air.

The wing is assumed to be a flat plate airfoil, where the $C_L$ and $C_D$ are given by *equations 8-9*. The wing is considered to have no twist i.e. $\gamma\lambda = 0$. The objective of the optimization is to obtain the chord function $c_\lambda(r)$ defined by *equation 13*, which is assumed to a polynomial of the five-order polynomial. Further increasing the polynomial order neither alters the wing shape nor improves the quality of the leading and trailing edge. The polynomial chord function ensures that the leading and trailing edge are smooth and continuous.

$$c_\lambda(r) = a_5 r^5 + a_4 r^4 + a_3 r^3 + a_2 r^2 + a_1 r^1 + a_0 r^0 \quad (13)$$

Moreover, to ensure chord at every $r$ does not reach impractical values, the chord function $c_\lambda(r)$ is bounded within $c_{max}$ and $c_{min}$, i.e. 0.01m and 0.04m respectively. Moreover, the length of wing is also constrained to $R = 0.25m$.

$$C = [c \in \mathbb{R} \mid c_{min} < c_\lambda(r) < c_{max}] \quad (14)$$

It is evident from the function that increasing the area of the wing increases the coefficient of power, however it also makes the overall pod heavier. Using inequality constraints, the wing area is constrained to have a reasonable weight.

$$A_{min} < \int_0^R c_\lambda(r)\, dr\, A_o < A_{max} \quad (15)$$

Further, for each wing's element, net moment is constrained to be zero. It is assumed that aerodynamic forces act at quarter of chord length from the leading edge of a wing's element. Whereas, the body forces like gravitational force are acting at the CG of the body.

$$M_{net} = \int M\, dr = 0 \quad (16)$$

With the constrained defined above, first the wing is divided into 1000 elements. Further, the value of axial induction factor (a) angular induction factor (a) and relative angle deviation (φ) are guessed for each wing's element by initializing the guessing algorithm with some arbitrary values of $a$ and $a'$ along with functions for the coefficient of lift and drag for a rectangular slab. The estimated values of $a$, $a'$, and $\varphi$ are then used in the objective function *equation 17* to derive the coefficient of power. The algorithm as shown below, is used to optimize for maximum coefficient of power $C_p$ using fmincon function in MATLAB optimization toolbox through Sequential Quadratic Programming (SQP) solver.

$$C_p(\gamma\lambda, c_\lambda(r), \varphi, a, a') = \frac{8}{\lambda_{max}^2} \int_{\lambda_{min}}^{\lambda_{max}} \lambda^3 J_\lambda\, d\lambda \quad (17)$$

Where, $J_\lambda$ is:

$$J_\lambda(\gamma\lambda, c_\lambda(r), \varphi, a, a') = F_\lambda(\varphi)\, a'\, (1-a)\left(1 - \frac{C_D(\varphi - \gamma\lambda)}{C_L(\varphi - \gamma\lambda)} \tan^{-1}\varphi\right) \quad (18)$$

---

**Algorithm:**
**Input:** Tol > 0, $\alpha \to C_L(\alpha), \alpha \to C_D(\alpha), \lambda, \gamma_\lambda, \sigma_\lambda, F_\lambda, x \to \psi(x)$
**Initial guess:** $a, a'$
**Output:** $a, a', \varphi$
Set err := Tol + 1
Define the function for $C_l$ and $C_d$ using equations 8 and 9
**while** err > Tol **do**
    Set $\varphi := \text{atan}\left(\frac{1-a}{\lambda(1+a')}\right)$
    Set $a := \frac{1}{\sin^2\varphi}(C_l(\varphi)\cos\varphi + C_d(\varphi)\sin\varphi)$
    Set $a' := \frac{1}{\lambda\sin^2\varphi}(C_l(\varphi)\sin\varphi - C_d(\varphi)\cos\varphi)$
    Set $err := \left|\tan\varphi - \frac{1-a}{\lambda(1-a')}\right|$
**end while**

---

The objective function and constraints were modified to obtain four different wing shapes, whose effects will be examined later.

- *W1* is the wing, where coefficient of power $C_p$ is maximized. However, flow over wing tip is ignored, hence the Prandtl Tip Function is assumed to be unity i.e. $F_\lambda(\varphi) = 1$. Moreover, the moment constraint is disregarded.
- *W2* is the wing, where coefficient of power $C_p$ is maximized. Similar to *W1*, flow over wing tip is ignored i.e. $F_\lambda(\varphi) = 1$, however the moment constraint is considered throughout the optimization.
- *W3* is the wing, where coefficient of power $C_p$ is maximized and flow over the wing tip is taken into account i.e. $F_\lambda(\varphi)$ is given by *equation 10*. Moreover, moment constraint is considered throughout the optimization.
- *W4* is an arbitrary rectangular wing of length $R$ and width of $c_{max}$. It acts as a control wing in simulation.

## 3. 6-DOF SIMULATION

Conducting a comprehensive six Degree of Freedom simulation is a necessity to verify and test the generated wing models. The simulation was done in MATLAB R2019B Simulink, using the 6-DOF Euler angles block. The calculation for the present state of the body is done by accessing the forces and moments incident on the body and then the calculation of the current state based on the values of the previous state. Here, two kinds of coordinate systems are used, one is fixed to the ground and is called the 'Earth Coordinate system' and the 'Body reference system'.

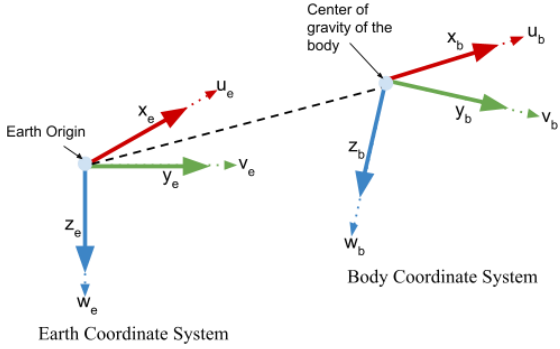

**FIGURE 4:** COORDINATE SYSTEM IN SIMULINK

The body's translatory motions are described by $F_b$, which contains the net forces in $x_b, y_b$, and $z_b$ directions on the body:

$$F_b = m(V_{b'} + \omega\, V_b) \qquad (19)$$

where $\omega$ are the rotation rates in the body frame of reference. The rotatory behavior is determined by the Moments which act on the body:

$$\bar{M}_B = \begin{bmatrix} L \\ M \\ N \end{bmatrix} = I\bar{\omega} + \bar{\omega} \times (I\bar{\omega}) \qquad (20)$$

$$I = \begin{bmatrix} I_{xx} & -I_{xy} & -I_{xz} \\ -I_{yx} & I_{yy} & -I_{yz} \\ -I_{zx} & -I_{zy} & I_{zz} \end{bmatrix} \qquad (21)$$

An element-based force and moment calculation was performed to determine the net force and moment on the body. The only force acting in the fixed direction w.r.t Earth Coordinate System is gravity, which means that it needs to be transformed to the Body Coordinate System to calculate its effect.

The expression for gravity force transformation for a body of mass $m$ is given by the following equations.

$$F_{g,e} = mg \qquad (22)$$

$$F_{g,b} = R \times F_{g,e} \qquad (23)$$

$$R = R_z(\psi)\, R_y(\theta)\, R_x(\phi) \qquad (24)$$

$$R = \begin{bmatrix} \cos(\psi) & -\sin(\psi) & 0 \\ \sin(\psi) & \cos(\psi) & 0 \\ 0 & 0 & 1 \end{bmatrix} \begin{bmatrix} \cos(\theta) & 0 & \sin(\theta) \\ 0 & 1 & 0 \\ -\sin(\theta) & 0 & \cos(\theta) \end{bmatrix}$$

$$\begin{bmatrix} 1 & 0 & 0 \\ 0 & \cos(\phi) & -\sin(\phi) \\ 0 & \sin(\phi) & \cos(\phi) \end{bmatrix} \qquad (25)$$

Where R is the rotational transformation matrix, dependent on the roll ($\phi$), pitch ($\theta$), and yaw ($\psi$) angle of the body w.r.t the fixed Earth Coordinate System.

The element forces and moments by the aerodynamics of the wing are primarily the net vertical lift acting in the negative $z_b$ direction, the net axial moment by the elemental drag around the $z_b$ axis, and the Lateral wing moment along the wing axis $x_b$. The vertical forces encountered by an individual element is given by:

$$F_{(z,b)} = \int_0^r \frac{1}{2} c_l\, v^2{}_{res}\, c(r)dr + \int_0^r \frac{1}{2} c_d\, v^2{}_{res}\, c(r)dr \qquad (26)$$

Where $c_l$ and $c_d$ are calculated at an effective angle of $(\theta - \phi_e)$ and $v^2{}_{res} = \omega_b{}^2 + (r\,\omega_{z,b})^2$. $c_\lambda(r)$ is the chord function determined by the chord optimization techniques. The net force vector can be expressed as:

$$F = F_{g,b} + \begin{bmatrix} 0 \\ 0 \\ -F_{(z,b)} \end{bmatrix} \qquad (27)$$

The net vertical forces are assumed to be centralized on the Centre of Gravity of the body due to the fact that these forces become radially symmetrical once reaching a high rotation rate and their net effect is only noticed along the $z_b$ axis. These forces work together against the gravitational force $F_{g,b}$ to produce lift in the body.

The net moment encountered by the body comes from two sources, the forces of drag through which the wing rotates around the $z_b$ axis and the individual moment of the wing element around the chord axis. The moment equations can be written as following:

$$M_{z,b} = \int_0^R \frac{1}{2} c_d\, v_{res}^2\, c(r)\, r\, dr - \int_0^R \frac{1}{2} c_l\, v_{res}^2\, c(r)\, r\, dr \qquad (28)$$

$$M_{x,b} = \int_0^r \frac{1}{2} c_m\, v_{res}^2\, c(r)\, r^2\, dr \qquad (29)$$

$$M = \begin{bmatrix} M_{x,b} \\ 0 \\ M_{z,b} \end{bmatrix} \qquad (30)$$

Where $c_l$ and $c_d$ are calculated at an effective angle of $(\theta - \phi_e)$ and $v^2{}_{res} = \omega_b{}^2 + (r\,\omega_{z,b})^2$. R is the maximum radial distance of the Wing and $c_\lambda(r)$ is the optimized chord function.

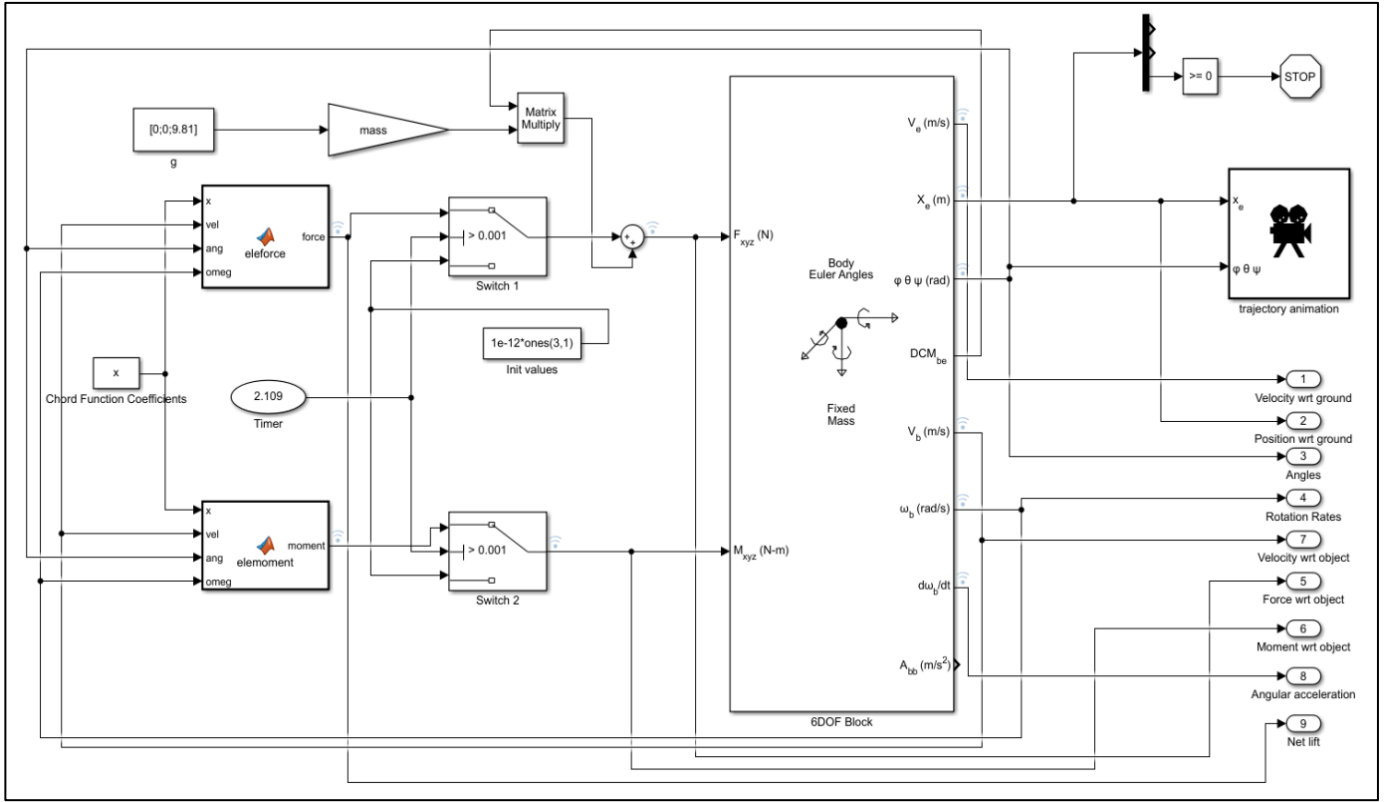

**FIGURE 5** SIMULINK MODEL CREATED FOR 6DOF SIMULATION OF THE SELECTED WING CANDIDATES. [7]

To setup the simulation in Simulink, an initialization quantity of *1e-12* is added to all the components of the Force and Moment for time less than *1e-3s* for numerical stability. The Simulink *6-DOF* [7] block takes in the $F_{xyz}$ and $M_{xyz}$ from the special functions `eleforce` and `elemoment`. These functions perform aforementioned numerical calculations on the wing profile which is defined by the variable '*x*'. '*x*' is the vector containing the coefficients of the $c_\lambda(r)$ polynomial defining the chord length w.r.t. the radial distance. Since this is a fixed mass simulation, the mass and inertia are previously given to the block. The functions `eleforce` and `elemoment` takes in the $V_b$, the roll ($\phi$), pitch ($\theta$), yaw ($\psi$) angles and $\omega_b$ of the current state to modify the forces and moments for the next state calculation. A separate block is required to convert the fixed gravitational acceleration to the dynamic coordinate system of the body. It is then added with the other forces together and fed into the '*6-DOF block*' [7]. These conditions are analogous to the conditions we are expecting it to be released. They system is assumed to working under standard atmospherics condition. The initial rest state is chosen to study the system states during transition from free fall to auto-rotation. The wing here, starts from the rest. The initial angle from which the wings are released is

$$[\phi_0 \quad \theta_0 \quad \psi_0] = \begin{bmatrix} 0 & \frac{\pi}{3} & 0 \end{bmatrix} \quad (31)$$

## 4. RESULT AND DISCUSSION

The coefficient of the chord function $c_\lambda(r)$ was optimized using MATALB `fmincon` function to maximize the coefficient of power $C_p$. Four different wings were obtained by varying the constraints and objective function, in order to evaluate the performance of ideal wing. Polynomial coefficients for W1, W2, W3 and W4 wings have been shown in *Figure 6*.

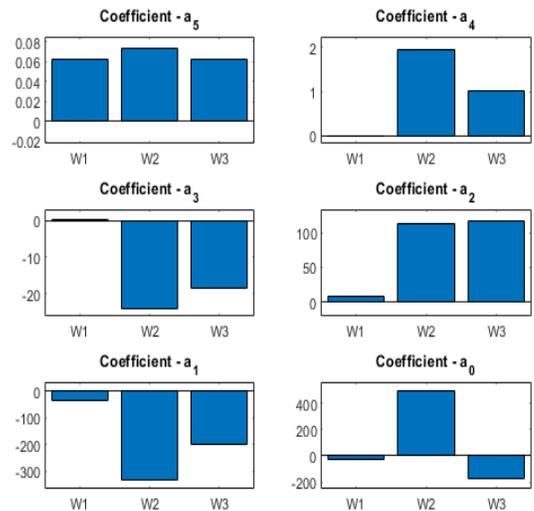

**FIGURE 6:** COEFFICIENTS OF CHORD FUNCTION FOR *W1*, *W2*, AND *W3*

| Parameter | Units | W1 | W2 | W3 | W4 |
|---|---|---|---|---|---|
| Wing Length | $R$ (m) | 0.25 | 0.25 | 0.25 | 0.25 |
| Wing Area | $A$ (m²) | 0.0179 | 0.0208 | 0.0208 | 0.0156 |
| Coefficient of Power | $C_P$ | 0.0140 | 0.0066 | 0.0072 | 0.0048 |
| Mass | $m$ (kg) | 0.1218 | 0.1229 | 0.1232 | 0.1207 |
| Moment of Inertia | $I$ (g.mm²) | $\begin{bmatrix} 5.79 & 2.85 & -4.28 \\ 2.84 & 58.7 & -0.32 \\ -4.27 & -0.32 & 59.5 \end{bmatrix} \cdot 1^{-5}$ | $\begin{bmatrix} 6.93 & 3.24 & -2.22 \\ 3.24 & 28.8 & -0.65 \\ -2.23 & -0.65 & 30.8 \end{bmatrix} \cdot 1^{-5}$ | $\begin{bmatrix} 6.08 & 2.98 & -2.64 \\ 2.98 & 39.5 & -0.39 \\ -2.64 & -0.39 & 40.9 \end{bmatrix} \cdot 1^{-5}$ | $\begin{bmatrix} 5.33 & 1.07 & -1.82 \\ 1.07 & 31.8 & -0.14 \\ -1.82 & -1.43 & 32.6 \end{bmatrix} \cdot 1^{-5}$ |
| System Conditions | | $\alpha = -0.15708 \, rad$ | | | |
| Environmental Conditions | | $\rho = 1.225 \, kg/m^3$, $\quad g = 9.81 m/s^2$ | | | |
| Initial Parameters for Pod | | $v_z = 0 \, m/s, \; \omega_z = 0 \, rad/s, z = 600m$ | | | |

**TABLE 1:** OPTIMIZED W1, W2, W3 AND W4 WING PARAMETERS AND SIMULATION PARAMETERS

The optimized chord function for *Wing 1* is denoted by *equation 32*, *Wing 2* is denoted by *equation 33*, and *Wing 3* is denoted by *equation 34*.

$$c_\lambda(r) = 0.063 \, r^5 - 4.52e^{-5} \, r^4 + 0.178 \, r^3 + 8.57 \, r^2 - 31.44 \, r - 22.76 \tag{32}$$

$$c_\lambda(r) = 0.074 \, r^5 + 1.94 \, r^4 - 23.89 \, r^3 + 112.82 \, r^2 - 328.44 \, r - 498.3 \tag{33}$$

$$c_\lambda(r) = 0.063 \, r^5 + 1.04 \, r^4 - 18.3 \, r^3 + 116.84 \, r^2 - 198.78 \, r - 175.53 \tag{34}$$

The *W1, W2, W3* and *W4* wings outline is shown in the *Figure 7*, where the vertical axes represent the non-dimensional chord and horizontal axis represents the distance of wing's elements from the root of the wing. Moreover, all the wings are aligned on the horizontal axis or $c_\lambda(r) = 0$. It represents the quarter chord location where the center of pressure is assumed for each element under steady state.

The coefficient of power $C_p$ for each optimized wing is shown in *table 1*. W1 has the highest coefficient of power, however the stability of the obtained wing is poor as the moment constraint was not taken into account. It is evident that *W3* has the highest coefficient of power $C_p$, which implies that during autorotation the pod with wing *W3* will have the slowest descent velocity. *W2* wing had a coefficient of power 8.3% lower than *W3*. Nevertheless, it is noted that both wings *W2* and *W3* had a coefficient of power $C_p$ higher than regular rectangular wing *W4*, about 27.2% and 33.3% higher respectively. W3 is the most ideal wing shape to have the most stable and lowest decent velocity in ideal conditions.

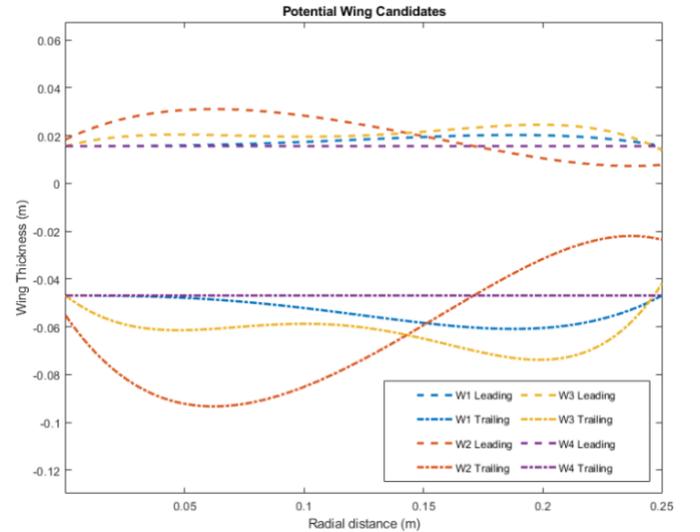

**FIGURE 7:** PLANFORM FOR OPTIMIZED WING *W1*, *W2*, *W3* AND *W4*

The optimized chord function for each wing were utilized to calculate the aerodynamic forces on the pod to calculate the $F_{xyz}$ and $M_{xyz}$ for 6-DOF simulation. Separates simulation were ran for each pod in the Simulink. From the *6-DOF* simulation data obtained from MATLAB R2019B Simulink with the initial conditions as explained earlier *equation 31*, the data on the velocities of the wing can be expressed with respect to time. The descent velocity $v_z$ for *W1, W2, W3* and *W4* are represented in *Figure 8*.

From the graph, it is evident that all of the wings show very stable velocities during the autorotation phase of the flight. Pod with *W2* wing shape, maintains at a vertical descent velocity of ~26m/s, which is quite high in comparison to other candidates. The one with the lowest descent velocity is Pod design with *W3* wing, of about 8.7 m/s. Moreover, the *x* and *y* velocities also become stable after some time in all of the four-pod model.

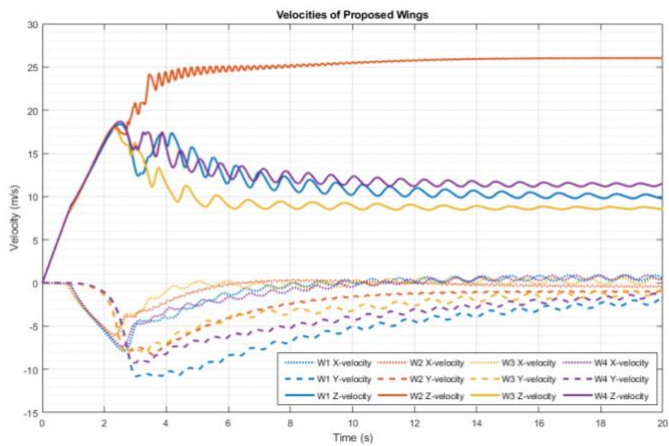

**FIGURE 8:** VELOCITY OF THE WINGS WITH RESPECT TO GROUND (m/s)

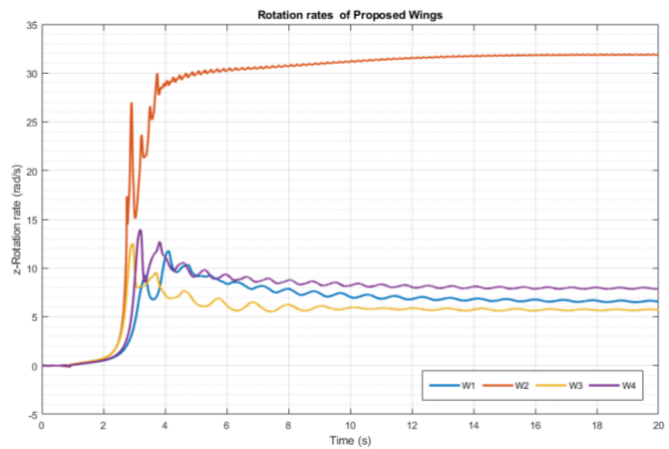

**FIGURE 9:** ROTATION RATE VARIATION OF THE WINGS (RAD/S)

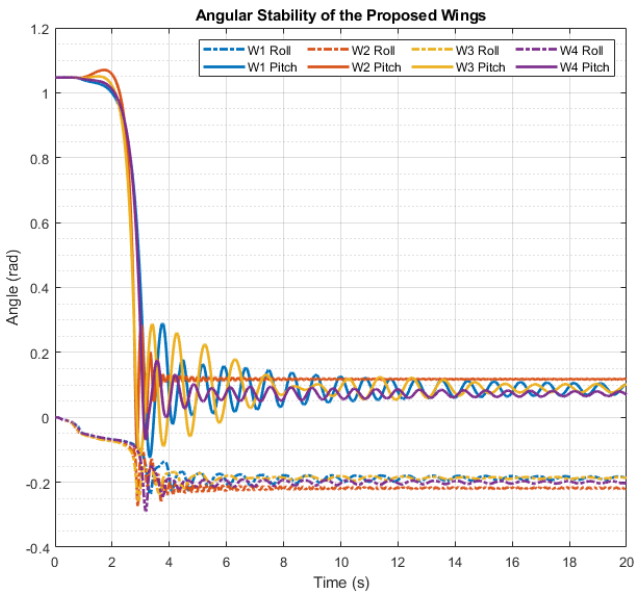

**FIGURE 10:** ANGULAR STABILITY OF WINGS

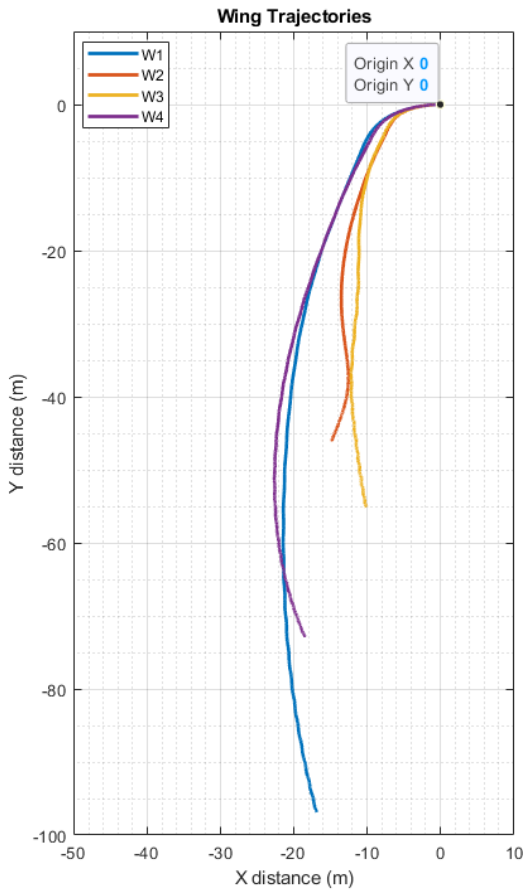

**FIGURE 11:** X-Y TRAJECTORIES IN IDEAL NO WIND CONDITIONS

    The Rotation rate of the wings around the *z axis* also attain a stable value after some time progression as represented in *Figure 9*. It is observed that pod with *W2* wing attains the highest rotation rate of ~32 rad/s whereas the other wings attain a stable value of ~5.8rad/s, ~7.0rad/s and ~8.1 rad/s for *W3, W1* and *W4* respectively. To further evaluate the stability of the systems while in autorotating phase of the flight, it is quintessential to examine the angular variation in *x axis* and *y axis* or the roll and pitch variation.

    From *Figure 10*, it is apparent that each wing starts off at the initial conditions as mentioned before and eventually when reaching autorotation, the roll and pitch angles stabilize within 3 second of the flight. Yaw is not mentioned here as the autorotation is happening around the z axis.

    The trajectory of each wing from the top after 20 seconds can be plotted as shown in *Figure 11*. As the wings reaches a tilt angle of about 60° around the *y axis*, the initial 'slide' towards the negative *x* direction is noticed. After achieving stable velocities in either direction, the mono-blade pod deviates from their path as shown in the *Figure 11*.

 

| Wings | Units | W1 | W2 | W3 | W4 |
|---|---|---|---|---|---|
| Total Mass (Kg) | $m$ (kg) | 0.1218 | 0.1229 | 0.1232 | 0.1207 |
| Stabilized Descent Velocity | $v_z$ (m/s) | 9.9520 | 26.0586 | 8.6336 | 11.3478 |
| Stabilized X – Velocity | $v_x$ (m/s) | 0.5295 | -0.4396 | 0.2539 | 0.6216 |
| Stabilized Y – Velocity | $v_y$ (m/s) | -2.1076 | -1.0098 | -1.2415 | -1.5842 |
| Rotation Rate | $\omega_z$ (rad/s) | 6.6025 | 31.9851 | 5.7074 | 7.8747 |
| Stable Roll Angle Achieved | rad | -0.1882 | -0.2183 | -0.1866 | -0.2016 |
| Stable Pitch Angle Achieved | rad | 0.0863 | 0.1117 | 0.0883 | 0.0744 |

**TABLE 2:** SIMULINK 6-DOF SIMULATION RESULTS SUMMARY

W3 which is moment stabilized by introducing moment constraint and tip loss corrections function during chordwise-optimization, performs significantly better than the rest of the candidates. Summary of the simulation for each pod concept are given in *Table 2*.

The analysis of the tradeoffs in *Table 2*, the following conclusions can be obtained: The involvement of the tip loss along with the moment balance along the wing axis while calculating the wing shape has dramatically affected the vertical descent rate and the rotation rate. *W3* shows this quite aptly. The wing shows the lowest vertical descent rate with the lowest rotation rate among all the test cases. *W2*, which is the wing where only moment balance is considered, does not show much promise regarding the vertical descent velocity; instead, it offers maximum stability. *Figure 10* shows the exponential decrease in the roll and pitch vibrations, which settles at a stable position faster than any of the other test cases. This wing shape shows the most promise in controllability. If control surfaces are added to this design, it will produce a much more stable system than any other. By disregarding the moment constraints and well the tip loss factor, the spatial stability is significantly affected. Both the wings *W1* and *W4* travel greater distances from the starting point on their own even when no external forces or any other stimuli were there to disturb the system.

## 5. CONCLUSION

In this letter, we presented an alternative method to design an unconventional descent method through Glauert's Model to optimize the shape of the wing. The mono-blade pod provides a silent and covert descent mechanism that requires no propulsive method. The proposed model performed close to biological maple seed. Moreover, the optimized wing resembles the shape of a maple seed. A Simulink model was successfully developed to recreate the non-linear flight model of the mono-blade pod. The 6-DOF simulation shows that the pod stabilizes itself due to autorotation, given that it is dropped at an angle. Through simulation and evaluating the coefficient of power, it is evident that the pod with a W3 wing had the ideal flight characteristics for a stable descent mechanism. W3 wing performed the best due to the inclusion of moment constraint and Prandtl Tip Function, which accurately models the flow across the tip.

To further validate our result, we can set up experimental experiments to verify the dynamic model and the flight characteristics of each wing under various real-life conditions. Future work consists of increasing the scale of the mono-blade pod concept to accommodate greater payload capacity. Further, various control methods could be implemented to guide the pod to a specific location or improve the flight's stability. The mono-blade pod concept could be used in agriculture and environmental survey tools due to its compact nature and the low manufacturing cost, where various sensors could be housed within the main body.